\documentclass[conference]{IEEEtran}
\IEEEoverridecommandlockouts
\usepackage{cite}
\usepackage{amsmath,amssymb,amsfonts}
\usepackage{algorithmic}
\usepackage{graphicx}
\usepackage{textcomp}
\usepackage{todonotes}
\usepackage{booktabs}
\usepackage{xcolor}
\usepackage{multirow}
\usepackage{tikz}
\usepackage{listings}
\usepackage{subfigure}

\definecolor{codegreen}{rgb}{0,0.6,0}
\definecolor{codegray}{rgb}{0.5,0.5,0.5}
\definecolor{codepurple}{rgb}{0.58,0,0.82}
\definecolor{backcolour}{rgb}{0.95,0.95,0.92}

\def\BibTeX{{\rm B\kern-.05em{\sc i\kern-.025em b}\kern-.08em
    T\kern-.1667em\lower.7ex\hbox{E}\kern-.125emX}}

\definecolor{backcolour}{rgb}{0.95,0.95,0.92}
\definecolor{codegreen}{rgb}{0,0.6,0}

\definecolor{codegreen}{rgb}{0,0.6,0}
\definecolor{codegray}{rgb}{0.5,0.5,0.5}
\definecolor{codepurple}{rgb}{0.58,0,0.82}
\definecolor{backcolour}{rgb}{0.95,0.95,0.92}
\definecolor{gray1}{HTML}{E6E6E6}

\definecolor{backcolour}{rgb}{0.95,0.95,0.92}
\definecolor{codegreen}{rgb}{0,0.6,0}

\definecolor{backcolour}{rgb}{0.95,0.95,0.92}
\definecolor{codegreen}{rgb}{0,0.6,0}

\lstdefinestyle{myStyle}{
    backgroundcolor=\color{gray1},   
    commentstyle=\color{codegreen},
    basicstyle=\ttfamily\footnotesize,
    breakindent=0em,
    breakatwhitespace=true,         
    breaklines=true,        
    columns=flexible,
    keepspaces=true,                 
    numbers=left,       
    numbersep=5pt,                  
    showspaces=false,                
    showstringspaces=false,
    showtabs=false,                  
    tabsize=2,
}

\usepackage{longtable}
\usepackage{framed}
\begin{document}

\title{
Game Generation via Large Language Models
}

\DeclareRobustCommand*{\IEEEauthorrefmark}[1]{%
    \raisebox{0pt}[0pt][0pt]{\textsuperscript{\footnotesize\ensuremath{#1}}}}

\author{\IEEEauthorblockN{Chengpeng Hu$^*$, Yunlong Zhao$^*$, Jialin Liu\thanks{Authors are with the Department of Computer Science and Engineering of the Southern University of Science and Technology, Shenzhen, China.}
\thanks{$^*$ Equal contribution.}
\thanks{This paper is accepted by 2024 IEEE Conference on Games.}}
}

 \maketitle

\begin{abstract}
Recently, the emergence of large language models (LLMs) has unlocked new opportunities for procedural content generation. However, recent attempts mainly focus on level generation for specific games with defined game rules such as \textit{Super Mario Bros.} and \textit{Zelda}. This paper investigates the game generation via LLMs. Based on video game description language, this paper proposes an LLM-based framework to generate game rules and levels simultaneously.
Experiments demonstrate how the framework works with prompts considering different combinations of context.
Our findings extend the current applications of LLMs and offer new insights for generating new games in the area of procedural content generation. 
\end{abstract}


\section{Introduction}

Recently, LLMs have been successfully applied to procedural content generation (PCG)~\cite{summerville2018procedural,liu2021deep,guzdial2022procedural,gallotta2024large}. Instead of generating game content via machine learning in the structured representation, LLMs involve inputting prompts, e.g. a short text, to generate game content.
The first ChatGPT for PCG Competition~\cite{Abdullah2024ChatGPT4PCG} was held at the 2023 IEEE Conference on Game, which aims at designing prompts that can generate specific \textit{Angry Bird} levels. The work of \cite{shyam2024mariogpt} combined LLMs with existing PCG methods and proposed MarioGPT to generate infinite \textit{Super Mario Bros.} levels via prompts. Todd \textit{et al.}~\cite{Todd2023level} investigated LLMs' capability in controllable level generation through a study on \textit{Sokoban}. Inspired by \cite{Todd2023level}, Nasir and Togelius~\cite{nasir2023practical} used GPT-3 to generate novel yet playable \textit{Dungeon}-like levels for a commercial game from a limited number of human-designed levels by data augmentation and bootstrapping. Despite of the success of LLMs in generating game levels, to the best of our knowledge, no existing work focuses on generating game rules and levels simultaneously.

This paper proposes an LLM-based framework to generate games (LLMGG), in which rule descriptions and levels are represented by video game description language (VGDL)~\cite{schaul2013video}. 
Several experiments with different combinations of prompts are conducted to verify LLMGG in generating game rules and levels simultaneously. This investigation helps to better understand the capabilities of LLMs and determines the potential benefits and limitations of generating games via LLMs.

\section{Preliminary}
\subsection{LLMs and Procedural Content Generation}
LLMs refer to the pre-trained language models with billions of parameters, aiming at solving general tasks~\cite{zhao2023survey}. Recent advancements such as GPTs~\cite{openai2023chatgpt}, Gemma~\cite{google2024gemma}, and Claude~\cite{anthropic2023claude} showcase their impressive ability in text generation, which demonstrates the potential of artificial general intelligence. As conversational agents, LLMs receive text-based prompts, understand human intentions, and generate corresponding answers via natural languages. 
This capability expands the boundary of LLMs more than mere text to other active domains such as game content generation~\cite{gallotta2024large}.

PCG considers to generate game content such as rules and levels~\cite{summerville2018procedural,liu2021deep}. Although tremendous progress has been made by combining PCG with machine learning, particularly deep learning, techniques~\cite{liu2021deep,guzdial2022procedural}, the trained models are limited to some specific game rules such as \textit{Super Mario Bros.} and thus lack generalisation. How these methods cooperate with humans to create game content is also challenging.
Natural language serves as an intuitive tool for communication among humans, while the emergence of LLMs enables cooperation between humans and machines through it.
Inputting text-based prompts is more straightforward than previously used inputs such as player trajectories for game content generation, thus allowing more involvement and feedback from human beings. 

Recent works have investigated into fine-tuning LLMs to generate levels given fixed game rules and datasets~\cite{shyam2024mariogpt,Todd2023level,nasir2023practical}. The tuned models enable humans to input some description, such as ``\textit{many coins, some pipes}", to create their own levels.
However, they require extra training time and are limited to the given games such as \textit{Super Mario Bros.}~\cite{shyam2024mariogpt} and \textit{Sokoban}~\cite{Todd2023level}. 
It is worth mentioning that GameGAN~\cite{kim2020learning} and Genie~\cite{bruce2024genie} generate the next game frame according to current frame and actions. The model itself can be considered as a game. However, the rules of generated games are not explicitly given to players. Moreover, how to modify the generated games by other algorithms or humans is not evident.
To the best of our knowledge, generating game rules, or both rules and levels simultaneously, via LLMs, has not yet been explored.
\begin{figure*}[ht]
    \centering
    \includegraphics[width=0.9\linewidth]{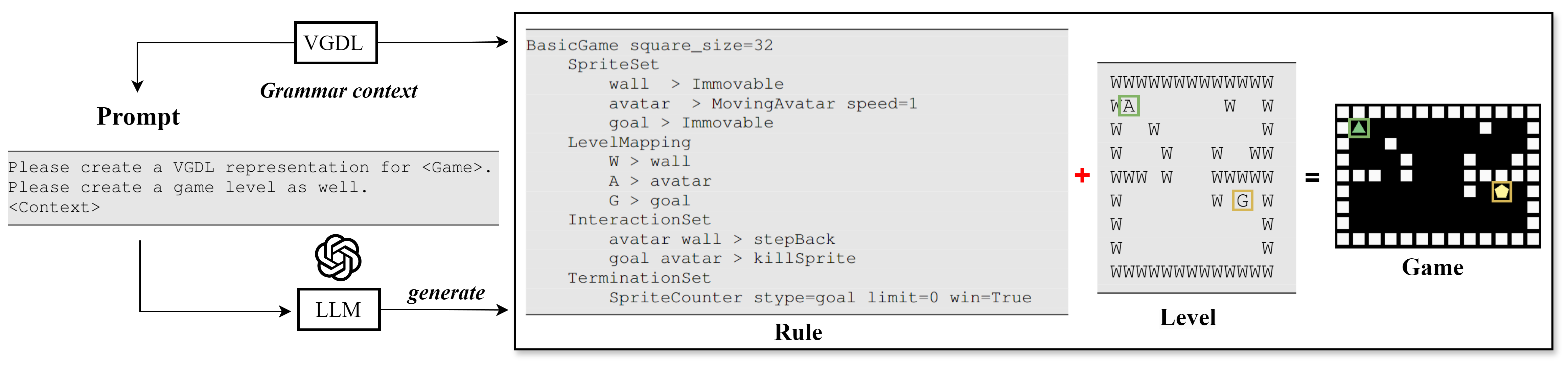}
    \caption{Framework of game generation via LLMs (LLMGG). An LLM receives a prompt to generate game rules and levels in the representation of VGDL.}
    \label{fig:wf_ggllm1}
\end{figure*}
\subsection{Video Game Description Language}
VGDL~\cite{schaul2013video} is a game description language framework for both game rules and levels, which has been used in general video game artificial intelligence (GVGAI)~\cite{khalifa2016general,perez2019general}. 
The right part of Fig. \ref{fig:wf_ggllm1} shows  a \textit{Maze} game based on VGDL.

There are four main components in VGDL, namely \textit{SpriteSet}, \textit{LevelMapping}, \textit{InteractionSet} and \textit{TerminationSet}. \textit{SpriteSet} defines sprites used in the game. \textit{LevelMapping} defines how those sprites are denoted and interact in levels. 
For example, ``\verb|avatar > MovingAvatar speed=1|" means there is a sprite named avatar, which is movable with property ``\verb|speed=1|". ``\verb|A > avatar|" means ``A'' denotes avatar, thus in the level file, ``A" indicates the place where an avatar will show up. 
\textit{InteractionSet} defines what happens if two sprites collide. ``\verb|avatar wall > stepBack|" means when an avatar collides with a wall, the avatar will step back. ``\verb|goal avatar > killSprite|" means when goal collides with an avatar, the goal will be removed. 
Taking \textit{Maze} as an example, the win rule requires the player to control an avatar to reach the goal.  
According to \textit{InteractionSet}, when a goal and avatar collide, the goal will be removed. Therefore, to make a winnable game, \textit{TerminationSet} keeps checking the number of goals. When the number of goals equals 0, i.e., ``\verb|limit=0|" the player wins and the game ends. Readers are referred to the work \cite{schaul2013video} for more about VGDL.

Humans can easily understand the mechanism of VGDL and rapidly design a game via natural language. Although machines also can directly read the description of game rules, it is sometimes hard to generate a good one via PCG methods due to the unstructured representation~\cite{nielsen2015towards,agarwal2023controllable}. There have been attempts to generate game levels and rules under the framework of VGDL~\cite{khalifa2017general,liebana2020general,bazzaz2023active}, however, they fail to directly incorporate human feedback.

\section{Generating Games via LLMs}
This section presents how our LLMGG framework (cf. Fig.~\ref{fig:wf_ggllm1}) generates games based on VGDL by inputting prompts. 

\subsection{Workflow of LLMGG}An LLM such as GPT first receives a text-based prompt.
The prompt contains some keywords such as VGDL and the expected game. Given the prompts, the LLM outputs consist of rules and levels in a text-based representation, which are then translated into an interactive game according to the syntax of VGDL by engines such as GVGAI gym~\cite{perez2019general}.

Notably, we do not assume any specific LLMs in this framework. It is believed that this framework provides a new insight as a general framework for game generation. Parts of the framework such as rule generation and level generation can 
interact iteratively and independently with LLMs for improving the prompt iteratively or fine-tuning LLMs.

\subsection{Prompt design}
\label{sec:pd}
Constructing effective prompts is crucial.
An example prompt is shown below:

\begin{lstlisting}[frame=bt,numbers=none]
Please create a VGDL representation for <Game>. 
Please create a game level as well.
<Context>
\end{lstlisting}
There are two options, namly \verb|<context>| and \verb|<Game>|. \verb|<context>| provides some explanations, e.g., VGDL grammar and examples of games encoded in VGDL of the prompt. \verb|<Game>| defines the expected game name by the users. Despite some common games like \textit{Maze}, known to LLMs, self-defined games are also supported, as long as the game mechanisms are described in \verb|<context>|.

\section{Generating \textit{Maze} via LLMs}
We generate \textit{Maze} games using our framework and discuss our observations. Experiments have been conducted on GPT-3.5, GPT-5 and Gemma 7B. \textbf{Supplementary Material} provides detailed examples of designed prompts and outputs. 

\subsection{Practical prompts for \textit{Maze}}

To determine the effect of \verb|<Context>|, we design seven prompts with different \verb|<Context>|, presented in Tab.~\ref{tab:pd_exp}. First, \verb|<Game>| is replaced by \verb|maze game|, noted as the basic instruction. ``Level'' defines how sprites map to characters like \verb|W| and \verb|A|. ``Base'' descriptions of grammar are directly copied from VGDL~\cite{schaul2013video}. $C_1$ and $C_2$ are type constraints such as sprite class and interaction methods. Specifically, the interaction method in $C_1$ is \verb|killSprite|, but \verb|removeSprite| in $C_2$. Besides, VGDL-based games with rules and levels extend the prompt, denoted as ``Example'' in Tab.~\ref{tab:pd_exp}.

\begin{table}[htbp]
    \centering
        \caption{Prompts designed under different contexts.}
    \begin{tabular}{c|c|c|c|c|c|c}
    \toprule
       \multirow{2}{*}{Prompt}  & \multirow{2}{*}{Instruction }&\multirow{2}{*}{Level}  & \multicolumn{3}{c|}{Grammar}& \multirow{2}{*}{Example} \\

        &  & & Base & $C_1$ & $C_2$\\
                \midrule
        $P_1$& \checkmark& & & & \\
        $P_2$& \checkmark&\checkmark & & & \\
        $P_3$& \checkmark& \checkmark& \checkmark& & \\
        $P_4$& \checkmark&\checkmark &\checkmark &\checkmark & \\
        $P_5$& \checkmark& \checkmark& \checkmark& & \checkmark\\
        $P_6$&  \checkmark&\checkmark &\checkmark & \checkmark& & \checkmark\\
        $P_7$& \checkmark&\checkmark & \checkmark& &\checkmark &\checkmark\\
        \bottomrule
    \end{tabular}

    \label{tab:pd_exp}
\end{table}

\begin{figure}[t]
    \centering
    \includegraphics[width=.85\linewidth]{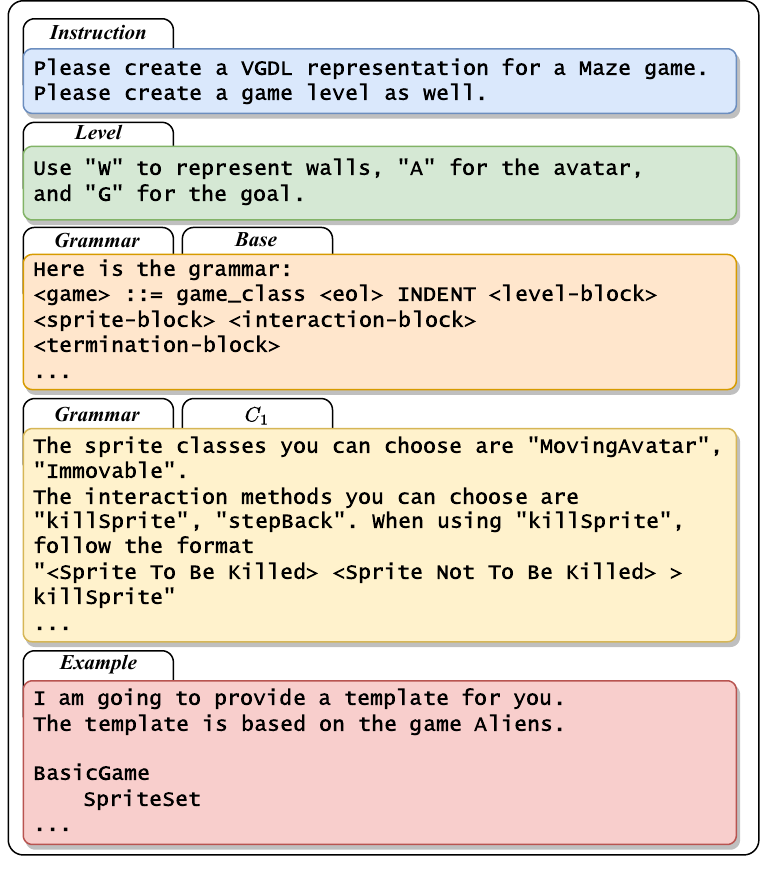}
    \caption{An example prompt of $P_6$, where ``$\cdots$" denotes omitted texts.}
    \label{fig:pe}
\end{figure}

\def\nouse{
An example prompt of $P_6$ is shown below:

\begin{lstlisting}
Please create a VGDL representation for a Maze game. 
Please create a game level as well.

Use ``W" to represent walls, ``A" for the avatar, 
and ``G" for the goal.
...
Here is the grammar:
<game> ::= game_class <eol> INDENT <level-block> 
<sprite-block> <interaction-block> 
<termination-block> 
...
The sprite classes you can choose are 
'MovingAvatar', "Immovable".
...
I am going to provide a template for you.
The template is based on the game Aliens.
```
BasicGame
    SpriteSet ...
```
 
\end{lstlisting}

}

\def\nouse{

Here is some fragment of the prompt:

\begin{itemize}
    \item LEVEL NOTATION: Use ``W" to represent walls, ``A" for the avatar, and ``G" for the goal.
    \item GRAMMAR: The grammar consists of two parts. The first one is the overview of VGDL. The second one is the grammar in Extended Backus-Naur Form. The The grammar is from \cite{schaul2013video}.
        \begin{lstlisting}[frame=bt,numbers=none]
...

The game description is composed of four blocks of instructions. 

The LevelMapping describes how to translate the characters in the level description into (one or more) objects, to generate the initial game state.
...

Here is the grammar:

<game> ::= game_class <eol> INDENT <level-block> <sprite-block> <interaction-block> <termination-block> 

<level-block> ::= LevelMapping <eol> INDENT { <char-map> NEWLINE } DEDENT 

...
        \end{lstlisting}
    \item Type Constraint 1:
        \begin{lstlisting}[frame=bt,numbers=none]
In the grammar, the sprite classes you can choose are 'MovingAvatar', 'Immovable'. 
        
The interaction methods you can choose are 'killSprite', 'stepBack'. When using 'killSprite', follow the format '<Sprite Be Killed> <Sprite Not Be Killed> > killSprite' .

The termination classes you can choose are 'SpriteCounter'. When using 'SpriteCounter', follow the format 'SpriteCounter stype=<Your Sprite Type> limit=<The number of your sprite type> win=<True/False>'.
        \end{lstlisting}

    \item Type Constraint 2:
        \begin{lstlisting}[frame=bt,numbers=none]
In the grammar, the sprite classes you can choose are 'MovingAvatar', 'Immovable'. 
        
The interaction methods you can choose are 'removeSprite', 'stepBack'. When using 'removeSprite', follow the format '<Sprite Not Be Removed> <Sprite Be Removed> > removeSprite'. 

The termination classes you can choose are 'SpriteCounter'. When using 'SpriteCounter', follow the format 'SpriteCounter stype=<Your Sprite Type> limit=<The number of your sprite type> win=<True/False>'.
        \end{lstlisting}

    \item Example:

            \begin{lstlisting}[frame=bt,numbers=none]
...

I am going to provide a template for you.
The template is based on the game Aliens.
```
BasicGame
    SpriteSet
        background > Immovable
        base    > Immovable    color=WHITE 
        avatar  > FlakAvatar   stype=sam
        missile > Missile

...

        avatar alien > killSprite scoreChange=-1
        avatar bomb  > killSprite scoreChange=-1
        alien  sam   > killSprite scoreChange=2
```

...
        \end{lstlisting}
\end{itemize}

The detailed prompt can be found in the Appendix.

}

\subsection{Rule-based text validation}

To validate the texts generated by LLMs, we set several language check rules: \textit{Parsable}, \textit{Logical} and \textit{Mappable}.
\textit{Parsable}: The generated VGDL should be recognised by a VGDL engine such as GVGAI. No grammar fault or keyword error is tolerated.
\textit{Logical}: According to VGDL, \textit{SpriteSet}, \textit{LevelMapping}, \textit{InteractionSet} and \textit{TerminationSet} are mandatory. Interaction completion ensures there are at least two interactions between avatar and wall, and between avatar and goal. Termination completion defines the win check, ensuring that a game should not terminate at the beginning and a win condition exists.
\textit{Mappable}: There should be correct mappings from characters to sprites. Characters with
``\#'' are prohibited.

\subsection{Necessity of context}

We conduct experiments with seven prompts designed following Tab.~\ref{tab:pd_exp}. Each is tested for 10 trials. Results are shown in Tab.~\ref{tab:metrics}.
We observe that only prompts with context generate correct games in terms of both rules and levels, e.g., GPT-3.5 and GPT-4 with $P_5$ and $P_7$. GPT-4 makes fewer mistakes than GPT-3.5, while Gemma 7B fails to generate parsable games almost all the trials. It is notable that only GPT-4 with $P_7$ outputs games with correct rules and levels in all the 10 trials.

Level mapping decides the conversion from natural language to digital content. There could be thousands of mappings that are available for LLMs. Our preliminary results show that, without predefined mapping (cf. $P_1$), LLM might present two different mapping in rules and levels, even though both of them make sense, independently. 
Although LLMs should understand VGDL syntax, which can be easily verified by asking LLMs about VGDL, however, without explicitly informing VGDL syntax in prompts, LLMs are more likely to generate imperfect VGDL text according to our experiments (cf. $P_1$ and $P_2$). Imperfections include lacking necessary components and incorrect indentation formats.

Extending the prompts with context (cf. $P_3$), some parsable VGDL texts are generated. This phenomenon might be related to in-context learning~\cite{dong2022survey}. Additional examples are also considered as contexts that demonstrate what the expected and correct output is. The incorporation of human knowledge helps LLMs to better understand the expectations of the instructions.

\begin{table}[t]
    \centering
        \caption{Result of generated games via different LLMs in 10 trials.}
    \begin{tabular}{lc|c|c|c|c}
\toprule

\multicolumn{2}{c|}{LLM}& \multicolumn{1}{c|}{Parsable}  & \multirow{1}{*}{Logical} &   \multicolumn{1}{c|}{Mappable} & \multicolumn{1}{c}{Correct}\\
\midrule
       \multirow{7}{*}{GPT-3.5}
        &$P_1$& 1 & 0 & 1 & 0\\
        &$P_2$& 2 & 0 & 3 & 0\\
        &$P_3$& 1 & 0 & 6 & 0\\
        &$P_4$& 10 & 0 & 7 & 0\\
        &$P_5$& 10 & 6  & 6 & 4\\
        &$P_6$& 10 & 0 & 8 & 0\\
        &$P_7$& 10 & 2  & 10 & 1\\
       \midrule
       \multirow{7}{*}{GPT-4}
        &$P_1$& 4 & 0 & 8& 0\\
        &$P_2$& 8 & 0 & 10& 0\\
        &$P_3$& 8 & 0 & 10& 0\\
        &$P_4$& 9 & 0 & 10& 0\\
        &$P_5$& 8 & 10 & 10 & 8\\
        &$P_6$& 10 & 0 & 9& 0\\
        &$P_7$& 10 & 10 & 10 & 10\\
         \midrule

       \multirow{7}{*}{Gemma 7B}
        &$P_1$& 0 & - & -& 0\\
        &$P_2$& 0 & - & -& 0\\
        &$P_3$& 0 & - & -& 0\\
        &$P_4$& 0 & - & -& 0\\
        &$P_5$& 0 &  - & -& 0\\
        &$P_6$& 4 & 4 & 2& 0\\
        &$P_7$& 0 &  - & -& 0\\
       \bottomrule
    \end{tabular}

    \label{tab:metrics}
\end{table}

\subsection{Wrong game logic due to LLM hallucination}

Although the levels converted from generated VGDL texts are parsable, they can be logically incorrect since game engines like GVGAI usually don't check the logic of interactions. There are two cases: \textit{incomplete interaction set} and \textit{wrong interaction sentence}. The former happens when the interaction only includes the logic of player and wall collision without describing the consequence of the collision. The later denotes possible logical errors. 
Usually, the avatar needs to touch a goal to win. When a goal and avatar collide, the goal should be removed, and the \textit{TerminationSet} detects if the number of goals is 0, and then the player wins the game. 
Therefore, the interaction should be written like ``\verb|goal avatar > killSprite|", in which \verb|killSprite| removes the first sprite (\verb|goal|). However, according to our experiments in Tab.~\ref{tab:pd_exp}, all the VGDL files generated by LLMs except with $P_5$ and $P_7$ set the interaction as ``\verb|avatar goal > killSprite|". 
We further ask LLMs to explain what ``\verb|avatar goal > killSprite|" means.
A typical explanation from LLMs is as follows:
\begin{lstlisting}[frame=bt,numbers=none,columns=fullflexible,breaklines=true]
"when the avatar reaches and collides with a goal sprite in the game, that particular goal sprite will be removed from the game board". 
\end{lstlisting}
However, the explanation is a wrong statement since the avatar will be removed actually instead of the goal.

\def\nouse{
Using prompt $P_3$ in ChatGPT 4.0, when the LLM generates the wrong logic ``goal avatar \textgreater ~killSprite", we keep asking ``explain avatar goal \textgreater ~killSprite within 15 words". The answer is ``When avatar contacts goal, the goal is removed, indicating objective completion or level success.". 
}
\def\nouse{
\begin{lstlisting}[frame=bt,numbers=none]
The answer is wrong, avatar goal > killSprite means
that when avatar contacts goal, the avatar is removed.
Please correct the VGDL representation.
\end{lstlisting}
}

\subsection{Hard to self-correct}
Based on the response illustrated above, we tell LLMs about the wrong statement and ask for a correction. The response below denotes that LLMs understand the rules after reminding, however, it does not how it works in programming.
\begin{lstlisting}[frame=bt,numbers=none]
"With the correction in mind, if we still intend for the avatar to reach the goal as a success condition rather than being removed from the game, we would need to adjust our understanding or the rules. However, based on your instruction, the rule indicates that the avatar is removed upon contacting the goal. This interaction might not directly reflect achieving the game's objective unless there's a specific design intention behind it, like a game mechanic where touching the goal triggers a victory condition indirectly."
\end{lstlisting}

\subsection{Avoiding hallucination by following LLMs?}

A new question arises: \textit{If we can not correct an LLM, why not following it?} 
Hence, a new interaction called ``\verb|removeSprite|" is created. It is similar to ``\verb|killSprite|". The only difference is that ``\verb|killSprite|" removes the first sprite, i.e., goal in the interaction \verb|goal avatar > killSprite|, while ``\verb|removeSprite|" removes the second sprite, i.e. avatar, in \verb|goal avatar > removeSprite|. It is observed from Tab.~\ref{tab:pd_exp}, only prompts with ``\verb|removeSprite|" (cf. $P_5$ and $P_7$) can generate correct games. Notably, all games generated in the 10 trials by GPT-4 with $P_7$ are correct.
A guess is that LLMs understand the word order in natural language, such as subject-verb-object or subject-object-verb. However, the syntax of ``\verb|killSprite|" in VGDL does not conform to the typical word order in natural language. LLMs leverage understanding of word order in natural language to the syntax of VGDL, which may lead to mistakes. However, it's hard to change the understanding of word order using only prompt engineering. Aligning the syntax to the understanding of LLMs is a more practical and feasible way.

\def\nouse{
\subsubsection{Necessity of repairing}
Tab. \ref{tab:errors} summarises errors the generated game may meet.
According to our empirical experiments, LLMs always generate some ''almost correct" games and levels. The generated rules and levels look similar to the true games that are preset in VGDL but retain some minor errors, which makes the generated games hardly recognised by the game engine.

}

\section{Conclusion}
This paper investigates the capability of LLMs in generating games. An LLM-based framework based on VDGL is proposed to generate game rules and levels simultaneously.
Taking \textit{Maze} as a case, several experiments are conducted on GPT-3.5, GPT-4 and Gemma. 
Results show the potential of using LLMs to generate game rules and levels by only modifying prompts.  
This enables human, particularly non-experts, to conveniently design their own game prototypes via LLMs. We also analyse the importance of contexts via designing different combinations of prompts and discuss the limitations of LLMs in hallucination and self-correct. 
Human intervention is still needed for directly generating complex and diverse games via LLMs.
In the future, we will extend the work to 3D game generation.

\bibliographystyle{IEEEtran} 
\bibliography{main}

\renewcommand{\appendixname}{Supplementary Material}

\newpage
\onecolumn
\appendix
\setcounter{figure}{0}
\setcounter{table}{0}

\subsection{Details of designed prompts}
The practical prompt consists of four parts including \textbf{Instruction}, \textbf{Level}, \textbf{Grammar} and \textbf{Example}, as explained in Sec.~\ref{sec:pd} of the main manuscript. By combining different parts, we get seven prompts, denoted in Tab.~\ref{tab:pd_exp}. Such combinations help examine the contribution of each part.
\subsubsection{Instruction prompt}
~~~~
\begin{lstlisting}[frame=bt,numbers=none]
Please create a VGDL representation for a maze game. Please create a game level as well.
\end{lstlisting}

\subsubsection{Level prompt}
~~~~

\begin{lstlisting}[frame=bt,numbers=none]
Use "W" to represent walls, "A" for the avatar, and "G" for the goal.
\end{lstlisting}

\subsubsection{Grammar}
There are three grammar prompts, namely Base, $C_1$ and $C_2$.
\paragraph{Base prompt}~~~~
\begin{lstlisting}[frame=bt, numbers=none]
A game is defined by two separate components, the level description, which essentially describes the positions of all objects and the layout of the game in 2D (i.e., the initial conditions), and the game description proper, which describes the dynamics and potential interactions of all the objects in the game.
The level description is simply a text string/file with a number of equal-length lines, where each character maps to (read: instantiates) one or more objects at the corresponding location of the rectangular grid.

The game description is composed of four blocks of instructions. 

The LevelMapping describes how to translate the characters in the level description into (one or more) objects, to generate the initial game state. For example, each "1" spawns an object of the "monster" class. 
The SpriteSet defines the classes of objects used, all of which are defined in the ontology, and derive from an abstract VGDLSprite class. Object classes are organized in a tree (using nested indentations), where a child class will inherit the properties of its ancestors. For example, there are two subclasses of avatars, one where Link possesses the key and one where he does not. Furthermore, all class definitions can be augmented by keyword options. For example, the "key" and "goal" classes differ only by their color and how they interact.

The InteractionSet defines the potential events that happen when two objects collide. Each such interaction maps two object classes to an event method (defined in the ontology), possibly augmented by keyword options. For example, swords kill monsters, monsters kill the avatar (both subclasses), nobody is allowed to pass through walls, and when Link finds a "key" object, the avatar class is transformed. 

The TerminationSet defines different ways by which the game can end, each line is a termination criterion, different criteria are available through the ontology and can be further specialized with keyword options. Here, it is sufficient to capture the goal (bring the sprite counter of the "goal" class to zero) to win.
What permits the descriptions to be so concise is an underlying ontology which defines many high-level building blocks for games, including the types of physics used (continuous, or grid based, friction, gravity, etc.), movement dynamics of objects (straight or random motion, player-control, etc.) and interaction effects upon object collisions (bouncing, destruction, spawning, transformation, etc.).

Here is the grammar:

<game> ::= game_class <eol> INDENT <level-block> <sprite-block> <interaction-block> <termination-block> 

<level-block> ::= LevelMapping <eol> INDENT { <char-map> NEWLINE } DEDENT 

<sprite-block> ::= SpriteSet <eol> INDENT { <sprite-def> NEWLINE } DEDENT 

<interaction-block> ::= InteractionSet <eol> INDENT { <interaction-def> <eol> } DEDENT 

<termination-block> ::= TerminationSet <eol> INDENT { <termination-def> <eol> } DEDENT 

<char-map> ::= CHAR " textgreater " <sprite-type> { " " <sprite-type> } 

<sprite-def> ::= <sprite-simple> [ <eol> INDENT { <sprite-def> <eol> } DEDENT ] 

<sprite-simple> ::= <sprite-type> " textgreater " [ sprite class ] { " " <option> } 

<interaction-def> ::= <sprite-type> <sprite-type> " textgreater " interaction method { " " <option> } 

<termination-def> ::= termination class { " " <option> } 

<eol> ::= { " " } [ "#" { CHAR | " " } ] NEWLINE 

<option> ::= IDENTIFIER "=" ( <sprite-type> | evaluable ) 

<sprite-type> ::= IDENTIFIER | "avatar" | "wall" | "EOS"
\end{lstlisting}

\paragraph{$C_1$ prompt}~~~~

\begin{lstlisting}[frame=bt,numbers=none]
In the grammar, the sprite classes you can choose are 'MovingAvatar', 'Immovable'. 
The interaction methods you can choose are 'killSprite', 'stepBack'. When using 'killSprite', follow the format '<Sprite To Be Killed> <Sprite Not To Be Killed> > killSprite' 
The termination classes you can choose are 'SpriteCounter'. When using 'SpriteCounter', follow the format 'SpriteCounter stype=<Your Sprite Type> limit=<The number of your sprite type> win=<True/False>'.
\end{lstlisting}
\paragraph{$C_2$ prompt}~~~~

\begin{lstlisting}[frame=bt,,numbers=none]
In the grammar, the sprite classes you can choose are 'MovingAvatar', 'Immovable'. 
The interaction methods you can choose are 'removeSprite', 'stepBack'. When using 'removeSprite', follow the format '<Sprite Not To Be Removed> <Sprite To Be Removed> > removeSprite'. 
The termination classes you can choose are 'SpriteCounter'. When using 'SpriteCounter', follow the format 'SpriteCounter stype=<Your Sprite Type> limit=<The number of your sprite type> win=<True/False>'.
\end{lstlisting}

\subsubsection{Example prompt} A classic game \textit{Aliens} is provided as the VGDL example.

\begin{lstlisting}[frame=bt,numbers=none]
I am going to provide a template for your output.
The template is based on the game Aliens.
```
BasicGame
    SpriteSet
        background > Immovable
        base    > Immovable    color=WHITE 
        avatar  > FlakAvatar   stype=sam
        missile > Missile
            sam  > orientation=UP    color=BLUE singleton=True 
            bomb > orientation=DOWN  color=RED  speed=0.5
        alien   > Bomber       stype=bomb   prob=0.01  cooldown=3 speed=0.8
            alienGreen > speed=0.8
            alienBlue > speed=0.8
        portal  > invisible=True
            portalSlow  > SpawnPoint   stype=alienBlue  cooldown=16   total=20
            portalFast  > SpawnPoint   stype=alienGreen  cooldown=12   total=20

    LevelMapping
        . > background
        0 > background base
        1 > background portalSlow
        2 > background portalFast
        A > background avatar

    TerminationSet
        SpriteCounter      stype=avatar               limit=0 win=False
        MultiSpriteCounter stype1=portal stype2=alien limit=0 win=True

    InteractionSet
        avatar  EOS  > stepBack
        alien   EOS  > turnAround
        missile EOS  > killSprite

        base bomb > killBoth
        base sam > killBoth scoreChange=1

        base   alien > killSprite
        avatar alien > killSprite scoreChange=-1
        avatar bomb  > killSprite scoreChange=-1
        alien  sam   > killSprite scoreChange=2
```
\end{lstlisting}

\clearpage
\subsection{Extended results}
Tab.~\ref{tab:errors} presents errors during generating games. In ``unparsable'' cases, there could be keyword and syntax errors. ``Component", ``Interaction" and ``Termination" are logical ``Illogical" errors.
``No level", ``Place", ``Mapping" and ``Sprite" are ``Unmappable" errors. They happen when mapping characters to sprites. ``Place" denotes that the level is in the wrong place of VGDL text and thus can not be recognised by the engine. ``Mapping" denotes wrong level mapping between rules and levels. ``Sprite" means that some vital sprites are missing in the generated level. It is easily observed from Tab.~\ref{tab:errors} that only GPT-4 with $P_7$ can generate games where both rules and levels are correct. However, Gemma 7B fails to generate either correct rules or levels for all prompts.

\begin{table*}[h]
    \centering
        \caption{Errors in generated rules and levels via LLMs among 10 trials. $G/R/L/W$ denotes that both rule and level are correct, correct rule but wrong level, wrong rule but correct level, and neither of them is correct.}
    \begin{tabular}{lc|c|c|c|c|c|c|c|c|c|c}
\toprule

\multicolumn{2}{c|}{\multirow{2}{*}{LLM}}& \multicolumn{2}{c|}{Unparsable}  & \multicolumn{3}{c|}{Illogical} & \multicolumn{4}{c|}{Unmappable} & \multirow{1}{*}{Correct}\\
&& Keyword&Syntax& Component& Interaction &Termination & No level & Place & Mapping & Sprite & $G/R/L/W$\\
\midrule
       \multirow{6}{*}{GPT-3.5}
         & $P_1$ &  6 & 4 & 9 & 3 & 1 & 0 & 4 & 9 & 1 &  0 / 0 / 1 / 9\\
         & $P_2$ &  2 & 6 & 7 & 6 & 1 & 0 & 1 & 4 & 2 &  0 / 0 / 4 / 6\\
         & $P_3$ &  3 & 5 & 0 & 10 & 2 & 0 & 3 & 3 & 0 & 0 / 0 / 7 / 3\\
         & $P_4$ &  0 & 0 & 0 & 10 & 2 & 0 & 0 & 3 & 0 & 0 / 0 / 7 / 3\\
         & $P_5$ &  0 & 0 & 0 & 4 & 0 & 0 & 0 & 2 & 2 &  \textbf{4} / 1 / 5 / 0\\
         & $P_6$ &  0 & 1 & 0 & 10 & 0 & 1 & 0 & 0 & 1 & 0 / 0 / 8 / 2\\
         & $P_7$ &  1 & 0 & 0 & 8 & 0 & 0 & 0 & 0 & 1 &  1 / 0 / 9 / 0\\
       \midrule
       \multirow{6}{*}{GPT-4}
         & $P_1$ & 3 & 4 & 0 & 10 & 0 & 0 & 0 & 1 & 1 & 0 / 0 / 8 / 2\\
         & $P_2$ & 1 & 1 & 1 & 10 & 0 & 0 & 0 & 0 & 0 & 0 / 0 / 10 / 0\\
         & $P_3$ & 1 & 1 & 0 & 10 & 0 & 0 & 0 & 0 & 0 & 0 / 0 / 10 / 0\\
         & $P_4$ & 0 & 1 & 0 & 10 & 0 & 0 & 0 & 0 & 0 & 0 / 0 / 10 / 0\\
         & $P_5$ & 1 & 1 & 0 & 10 & 0 & 0 & 0 & 0 & 0 & 8 / 0 / 2 / 0\\
         & $P_6$ & 0 & 0 & 0 & 10 & 0 & 0 & 1 & 0 & 0 & 0 / 0 / 9 / 1\\
         & $P_7$ & 0 & 0 & 0 & 0 & 0 & 0 & 0 & 0 & 0 & \textbf{10} / 0 / 0 / 0\\
         \midrule

       \multirow{6}{*}{Gemma 7B}
         & $P_1$ & 0 & 10 & - & - & - & - & - & - & - & 0 / 0 / 0 / 10\\
         & $P_2$ & 0 & 10 & - & - & - & - & - & - & - & 0 / 0 / 0 / 10\\
         & $P_3$ & 0 & 10 & - & - & - & - & - & - & - & 0 / 0 / 0 / 10\\
         & $P_4$ & 0 & 10 & - & - & - & - & - & - & - & 0 / 0 / 0 / 10\\
         & $P_5$ & 0 & 10 & 0 & - & - & 0 & 0 & 0 & 0 & 0 / 0 / 0 / 10\\
         & $P_6$ & 4 & 2 & 0 & 6 & 6 & 0 & 1 & 0 & 0 & 0 / 0 / 0 / 10\\
         & $P_7$ & 10 & 3 & 0 & - & - & 4 & 0 & 1 & 1 & 0 / 0 / 0 / 10\\
       \bottomrule
    \end{tabular}

    \label{tab:errors}
\end{table*}

\definecolor{mygreen}{rgb}{0,0.6,0}
\definecolor{mygray}{rgb}{0.5,0.5,0.5}
\definecolor{mymauve}{rgb}{0.58,0,0.82}

\subsection{Outputs of LLMs}
The outputs of LLMs including GPT-3.5, GPT-4 and Gemma 7B in terms of $P_1$, $P_2$, $P_3$, $P_4$, $P_5$, $P_6$ and $P_7$ are summarised in this section.

\subsubsection{Output of GPT-3.5}
Seven representative outputs of GPT-3.5 via seven prompts are present below.
\paragraph{Output of GPT-3.5 via \textbf{$P_1$}}
~~~~
\begin{lstlisting}[frame=bt,numbers=none]
VGDL (Video Game Description Language) is a domain-specific language used for describing 2D video games. Here's a simple VGDL representation for a maze game along with a sample game level:

```python
BasicGame
    SpriteSet
        wall > Immovable
        player > MovingAvatar
        goal > Immovable
    LevelMapping
        # represents the game level layout
        0 > wall
        1 > player
        2 > goal
    TerminationSet
        SpriteCounter stype=player limit=0 win=True
        SpriteCounter stype=goal limit=0 win=False
```

This VGDL description defines a basic maze game with three types of sprites: wall, player, and goal. The player can move around the maze, the goal is where the player must reach, and walls are obstacles that cannot be traversed. The game terminates when either the player sprite is eliminated (indicating the player lost) or when all goal sprites are eliminated (indicating the player won).

Here's a simple game level represented using numbers where 0 represents walls, 1 represents the player's starting position, and 2 represents the goal:

```
1111111111
1000000001
1010101011
1000000001
1010101011
1000000001
1010101011
1000000001
1010101011
1000000001
1111111111
```

This level creates a maze where the player (1) starts at the top left corner and must navigate to the goal (2) at the bottom right corner while avoiding walls (0).
\end{lstlisting}

\paragraph{Output of GPT-3.5 via \textbf{$P_2$}}
~~~~
\begin{lstlisting}[frame=bt,numbers=none]
VGDL (Video Game Description Language) is a framework for describing 2D video games. Below is a VGDL representation for a simple maze game, along with a sample game level.

```plain
BasicGame
    SpriteSet
        wall > Immovable
            color=GREY
        avatar > MovingAvatar
            speed=1
            cooldown=0
            color=BLUE
        goal > Immovable
            color=YELLOW
    TerminationSet
        SpriteCounter stype=avatar limit=0 win=True
    LevelMapping
        W=wall
        A=avatar
        G=goal

```

This VGDL code defines a simple maze game with an avatar (player), walls, and a goal. The objective is for the avatar to reach the goal while navigating through the maze.

Now, let's create a sample game level using the provided VGDL representation:

```plain
WWWWWWWWWWWWWW
W      W     W
W  A       G W
W      W     W
WWWWWWWWWWWWWW
```

In this game level:

- "W" represents walls, forming a maze structure.
- "A" represents the avatar, starting position of the player.
- "G" represents the goal, where the player needs to reach.
- Empty spaces represent open areas where the player can move.

You can adjust the maze layout and size as desired within the VGDL framework.
\end{lstlisting}
\clearpage
\paragraph{Output of GPT-3.5 via \textbf{$P_3$}}
~~~~
\begin{lstlisting}[frame=bt,numbers=none]
Here's a VGDL representation for a maze game along with a sample game level:

```vgdl
BasicGame
    LevelMapping
        W > wall
        A > avatar
        G > goal

    SpriteSet
        avatar > MovingAvatar
        wall > Immovable
        goal > Immovable

    InteractionSet
        avatar wall > stepBack
        avatar goal > killSprite

    TerminationSet
        SpriteCounter stype=goal limit=0

```

Sample Game Level:
```
WWWWWWWWWWWWWWWW
W              W
W   A        G W
W              W
W              W
W              W
WWWWWWWWWWWWWWWW
```

In this representation:

- **LevelMapping**: Describes how characters in the level file translate into objects in the game. Here, 'W' represents walls, 'A' represents the avatar, and 'G' represents the goal.
  
- **SpriteSet**: Defines the classes of objects used in the game. In this case, we have the avatar (MovingAvatar), walls (Immovable), and the goal (Immovable).
  
- **InteractionSet**: Describes potential interactions between objects. The avatar interacts with walls by stepping back when colliding with them and kills the goal when reaching it.
  
- **TerminationSet**: Defines how the game can end. In this case, the game ends when the sprite counter for the goal reaches zero, indicating the goal has been reached.
\end{lstlisting}
\paragraph{Output of GPT-3.5 via \textbf{$P_4$}}
~~~~
\begin{lstlisting}[frame=bt,numbers=none]
Here's a VGDL representation for a maze game along with a sample game level:

```plaintext
BasicGame
    LevelMapping
        W > wall
        A > MovingAvatar
        G > goal
    SpriteSet
        wall > Immovable
        MovingAvatar > MovingAvatar
        goal > Immovable color=GREEN
    InteractionSet
        MovingAvatar wall > stepBack
        MovingAvatar goal > killSprite
    TerminationSet
        SpriteCounter stype=goal limit=0 win=True
```

Sample game level:

```plaintext
#######
#A# #G#
# # # #
# #   #
#   # #
####### 
```

In this representation:

- The avatar (A) is controlled by the player and can move in the maze.
- Walls (W) are immovable objects that block the avatar's path.
- The goal (G) is the destination the avatar needs to reach to win the game.
- When the avatar collides with a wall, it steps back.
- When the avatar reaches the goal, it is removed from the game.
- The game is won when the counter of goal sprites reaches zero.
\end{lstlisting}
\paragraph{Output of GPT-3.5 via \textbf{$P_5$}}
~~~~
\begin{lstlisting}[frame=bt,numbers=none]
Here's a VGDL representation for a maze game along with a simple game level:

```plaintext
BasicGame
    LevelMapping
        W > Immovable
        A > MovingAvatar
        G > Immovable
    SpriteSet
        MovingAvatar > avatar
        Immovable > wall
    InteractionSet
        avatar wall > stepBack
        avatar wall > removeSprite
        avatar goal > removeSprite
    TerminationSet
        SpriteCounter stype=goal limit=0 win=True
```

And a sample game level:

```plaintext
#######
#A  W #
# # # #
#   W #
# # # #
# G   #
#######
```

In this representation:

- `LevelMapping` maps the characters in the level description to sprite classes.
- `SpriteSet` defines the classes of objects used, such as the avatar and walls.
- `InteractionSet` specifies what happens when certain objects collide. Here, the avatar interacts with walls and the goal.
- `TerminationSet` defines the condition for winning the game, which is when the goal sprite counter reaches zero.

The sample game level consists of walls represented by "W", the avatar represented by "A", and the goal represented by "G".
\end{lstlisting}
\clearpage
\paragraph{Output of GPT-3.5 via \textbf{$P_6$}}
~~~~
\begin{lstlisting}[frame=bt,numbers=none]
```
BasicGame
    SpriteSet
        wall   > Immovable
        avatar > MovingAvatar    speed=0.1
        goal   > Immovable
    LevelMapping
        W > wall
        A > avatar
        G > goal

    TerminationSet
        SpriteCounter stype=goal limit=0 win=True

    InteractionSet
        avatar wall > stepBack
        avatar goal > killSprite
```

And here's a sample level for the maze game:

```
WWWWWWWWWW
WAG      GW
W         W
W         W
W         W
WWWWWWWWWW
```
\end{lstlisting}
\paragraph{Output of GPT-3.5 via \textbf{$P_7$}}
~~~~
\begin{lstlisting}[frame=bt,numbers=none]
Here's the VGDL representation for a maze game along with a sample game level:

```plaintext
BasicGame
    SpriteSet
        wall    > Immovable
        avatar  > MovingAvatar
        
    LevelMapping
        W > wall
        A > avatar
        G > goal

    TerminationSet
        SpriteCounter stype=goal limit=0 win=True

    InteractionSet
        avatar wall > stepBack
```

Sample game level:

```
WWWWWWWWWW
WAG      W
W   WWW  W
W   WGW  W
W   WWW  W
W        W
WWWWWWWWWW
```

In this game, the avatar ('A') navigates through the maze to reach the goal ('G'), while avoiding the walls ('W'). The goal of the game is to reach the goal position.
\end{lstlisting}

\clearpage
\subsubsection{Output of GPT-4}
Seven representative outputs of GPT-4 via seven prompts are present below.

\paragraph{Output of GPT-4 via \textbf{$P_1$}}
~~~
\begin{lstlisting}[frame=bt,numbers=none]
Creating a Video Game Description Language (VGDL) representation for a maze game involves defining the game's rules, mechanics, and the level layout in a structured format that can be understood by a game engine or framework designed to interpret VGDL scripts. Let's start by defining the VGDL for a simple maze game where the player's objective is to navigate through a maze to reach a goal. In this game, walls block the player's path, and the player needs to find a way to the goal avoiding dead ends.

### VGDL Game Description

The VGDL game description consists of two parts: the game definition and the level map. First, we'll outline the game definition, which includes the game's sprite set, interaction set, and termination conditions.

#### Sprite Set
Defines the game objects.
```plaintext
SpriteSet
  avatar > MovingAvatar
  goal > Immovable
    winGoal > goal
  wall > Immovable
```

#### Interaction Set
Defines how game objects interact with each other.
```plaintext
InteractionSet
  avatar wall > stepBack  # The player cannot pass through walls
  avatar winGoal > killSprite  # When the player reaches the goal, the level is won
```

#### Termination Conditions
Defines the conditions under which the game ends.
```plaintext
TerminationSet
  SpriteCounter stype=winGoal limit=0 win=True  # Win when all winGoals are reached
```

#### Level Mapping
Defines the characters used in the level map to represent game objects.
```plaintext
LevelMapping
  G > winGoal
  W > wall
  A > avatar
```

### Level Design

Now, let's create a simple level for our maze game. In the level map, 'W' represents walls, 'A' the player's starting position, and 'G' the goal.

```plaintext
Level
WWWWWWWWWW
WA      GW
W WWWWW WW
W        W
WWWWWWWWWW
```

In this level, the player (`A`) starts near the bottom left corner and needs to navigate through the maze to reach the goal (`G`) at the far right.

This basic VGDL representation and level design for a maze game provides a straightforward maze exploration challenge. You can extend this by adding more complex mazes, additional game mechanics (like keys to unlock doors, enemies, or time limits), and varied level designs to enhance the gameplay experience. VGDL is flexible and allows for a wide range of game types and mechanics to be described and implemented.
\end{lstlisting}

\clearpage
\paragraph{Output of GPT-4 via \textbf{$P_2$}}

~~~
\begin{lstlisting}[frame=bt,numbers=none]
To design a Video Game Description Language (VGDL) representation for a simple maze game along with a game level, we'll follow the basic structure of VGDL, which involves defining the game's sprite set, level mapping, and the interactions. VGDL is used in research and education to quickly prototype and study games.

### VGDL Representation of the Maze Game

In a VGDL file, the game is typically divided into several sections: `SpriteSet`, `LevelMapping`, `InteractionSet`, and `TerminationSet`. Here's a simple example:

#### SpriteSet
Defines the game objects.

```
SpriteSet
    Wall > Immovable
    Avatar > MovingAvatar
    Goal > Passive
```

#### LevelMapping
Defines what characters in the level map to which sprites.

```
LevelMapping
    W > Wall
    A > Avatar
    G > Goal
```

#### InteractionSet
Defines the interactions between objects. For a simple maze game, we might have the avatar win upon reaching the goal, and nothing happens upon touching walls.

```
InteractionSet
    Avatar wall > stepBack
    Avatar goal > win
```

#### TerminationSet
Defines the conditions under which the game can end.

```
TerminationSet
    Goal collected > win
```

### Example Game Level

For the game level, you use a grid of characters where "W" represents walls, "A" the avatar, and "G" the goal. Here's a simple level example:

```
WWWWWWWWWW
W        W
W W WWW  W
W W  A   W
W WWWWW WW
W        W
W WWWWWWWW
W       GW
WWWWWWWWWW
```

This level structure creates a maze with walls ("W"), an avatar starting position ("A"), and a goal ("G"). The avatar must navigate through the maze to reach the goal.

Combining these elements, you would have a basic VGDL description and a level for a simple maze game. To implement this game, you would use a VGDL parser or engine that can interpret these definitions and run the game accordingly.
\end{lstlisting}
\clearpage
\paragraph{Output of GPT-4 via \textbf{$P_3$}}
~~~
\begin{lstlisting}[frame=bt,numbers=none]
Given the guidelines for VGDL, let's create a simple maze game. This game will feature an avatar that the player controls, walls that form the maze, and a goal the player aims to reach. The level design will reflect these elements, ensuring there are paths for the avatar to navigate through the maze to reach the goal.

### VGDL Game Description

**Game Description (MazeGame.vgdl):**
```
BasicGame
    LevelMapping
        W > wall
        A > avatar
        G > goal
    SpriteSet
        wall > Immovable color=GRAY
        avatar > MovingAvatar
        goal > Passive color=GOLD
    InteractionSet
        avatar wall > stepBack
        avatar goal > killSprite
    TerminationSet
        SpriteCounter stype=goal limit=0 win=True
```

**Explanation:**
- **LevelMapping:** Maps the characters 'W', 'A', and 'G' to wall, avatar, and goal sprites, respectively.
- **SpriteSet:** Defines the objects in the game. Walls are immovable and gray, representing the maze's barriers. The avatar is a moving sprite controlled by the player. The goal is a passive, gold-colored object the player aims to reach.
- **InteractionSet:** Defines interactions between sprites. If the avatar collides with a wall, it must step back, preventing it from passing through walls. When the avatar reaches the goal, the goal sprite is removed from the game.
- **TerminationSet:** The game ends when the number of goal sprites reaches zero, indicating the player has reached the goal, winning the game.

### Level Design

**Level Layout (MazeLevel.txt):**
```
WWWWWWWWWWWWWWWW
WA            GW
W WWWWWWWW WWWWW
W               W
WWWW WWWWW WWWWW
W               W
W WWWWWWW WWWWW W
W               W
WWWWWWWWWWWWWWWW
```

**Explanation:**
- **W:** Represents walls that form the boundaries and internal maze structure.
- **A:** The starting position of the avatar.
- **G:** The goal's position, which the avatar aims to reach.

This simple maze game and level design follow the VGDL principles, providing a clear distinction between static and interactive elements within the game. The layout ensures there's a clear objective, with challenges presented by the maze structure that the player must navigate through.
\end{lstlisting}
\clearpage
\paragraph{Output of GPT-4 via \textbf{$P_4$}}
~~~~
\begin{lstlisting}[frame=bt,numbers=none]
Let's create a simple VGDL representation for a maze game along with a game level. In this example, the maze will have walls, an avatar the player controls, and a goal to reach. The VGDL game description will include four main blocks as per the provided grammar: LevelMapping, SpriteSet, InteractionSet, and TerminationSet. Here is how these components can be assembled:

### VGDL Game Description

```yaml
BasicGame 
    LevelMapping 
        W > wall 
        A > avatar 
        G > goal
    SpriteSet 
        wall > Immovable 
        avatar > MovingAvatar 
        goal > Immovable
    InteractionSet 
        avatar wall > stepBack 
        avatar goal > killSprite 
    TerminationSet 
        SpriteCounter stype=goal limit=0 win=True
```

This game description outlines the basic components of our maze game. The `LevelMapping` section defines the characters for walls, the avatar, and the goal. The `SpriteSet` section classifies these characters into sprite classes, with walls and the goal being immovable objects, while the avatar is a moving object that the player controls. The `InteractionSet` specifies that the avatar cannot pass through walls (it will step back upon collision), and touching the goal will "kill" the goal sprite, which is how the player wins the game. The `TerminationSet` defines the win condition as eliminating all goal sprites from the level.

### Game Level

Here is a simple level for this game. The level layout uses "W" for walls, "A" for the avatar's starting position, and "G" for the goal. The level is a simple maze that the player must navigate through to reach the goal.

```
WWWWWWWWWW
WA      GW
W WWWW WWW
W        W
WWWWWWWWWW
```

In this level, the avatar starts near the top-left corner and must navigate through the maze to reach the goal on the right side. Walls are placed to create a path that the avatar must follow to reach the goal. This layout can be easily modified to create more complex mazes or different game experiences.
\end{lstlisting}
\clearpage
\paragraph{Output of GPT-4 via \textbf{$P_5$}}
~~~
\begin{lstlisting}[frame=bt,numbers=none]
Let's create a simple VGDL description for a basic maze game along with a level layout that follows the rules you've provided.

### VGDL Game Description

**BasicGame**
```
LevelMapping
    W > wall
    A > avatar
    G > goal

SpriteSet
    wall > Immovable
    avatar > MovingAvatar
    goal > Immovable

InteractionSet
    avatar wall > stepBack
    avatar goal > removeSprite goal

TerminationSet
    SpriteCounter stype=goal limit=0 win=True
```

This VGDL game description defines a simple maze game where the player's objective is to reach the goal. The player controls the `avatar`, which can move around but must avoid `walls`. Upon reaching the `goal`, the game ends in victory.

### Level Description

For the level layout, let's create a small maze. The layout uses "W" for walls, "A" for the avatar's starting position, and "G" for the goal. Here's an example level:

```
WWWWWWWWWW
WA      GW
W W WWW  W
W W  W   W
W WWWWW WW
W        W
WWWWWWWWWW
```

In this level, the avatar starts near the left side of the maze and must navigate through the openings in the walls to reach the goal on the opposite side. The maze is surrounded by walls on all sides, ensuring the player stays within the bounds of the maze.
\end{lstlisting}
\clearpage
\paragraph{Output of GPT-4 via \textbf{$P_6$}}
~~~~
\begin{lstlisting}[frame=bt,numbers=none]
Based on your description, I'll create a VGDL representation for a simple maze game, adhering to the structure and constraints provided. This maze game will include walls (W), an avatar (A), and a goal (G). Here's how the game and a basic level might be defined:

```yaml
BasicGame
    SpriteSet
        wall    > Immovable    color=GRAY
        avatar  > MovingAvatar color=GREEN
        goal    > Immovable    color=GOLD singleton=True

    LevelMapping
        W > wall
        A > avatar
        G > goal

    InteractionSet
        avatar wall > stepBack
        avatar goal > killSprite # When the avatar reaches the goal, the game is won

    TerminationSet
        SpriteCounter stype=goal limit=0 win=True
```

And a simple level for this game could look like:

```
WWWWWWWWWWWW
W        W G
W WWWWWW W W
W W    W W W
W W WW W W W
W W WW W W W
A W    W   W
WWWWWWWWWWWW
```

In this level description:
- `W` represents walls that the avatar cannot pass through.
- `A` marks the starting position of the avatar.
- `G` is the goal that the avatar needs to reach to win the game.

The game mechanics are simple:
- The avatar can move in four directions (up, down, left, right).
- It cannot move through walls (`W`), and attempting to do so will result in the avatar stepping back to its previous position.
- The game is won when the avatar reaches the goal (`G`).
- The goal is a singleton object, meaning there is only one goal in the level, and reaching it triggers the winning condition through the interaction and termination sets defined.
\end{lstlisting}
\clearpage
\paragraph{Output of GPT-4 via \textbf{$P_7$}}
~~~~
\begin{lstlisting}[frame=bt,numbers=none]
Given the rules and the structure provided by the VGDL grammar, we will design a simple maze game. The objective of this game is for the player (Avatar) to navigate through a maze of walls to reach a goal. Here is how it could be structured based on your requirements:

### VGDL Game Description

```vgdl
BasicGame
    SpriteSet
        wall    > Immovable    color=GRAY
        avatar  > MovingAvatar color=BLUE
        goal    > Immovable    color=GOLD singleton=True

    LevelMapping
        W > wall
        A > avatar
        G > goal

    TerminationSet
        SpriteCounter      stype=goal    limit=0 win=True

    InteractionSet
        avatar wall > stepBack
        avatar goal > removeSprite
```

### Level Description

For the level, we'll create a simple maze where "W" represents walls, "A" for the avatar's starting position, and "G" for the goal. The level is designed as a 9x9 grid for this example. You can modify this grid size according to your needs or game complexity.

```level
WWWWWWWWW
WA      W
W W WWW W
W W    GW
W WWWWW W
W       W
WWWWWWWWW
```

This level sets up a simple maze where the avatar must navigate from its starting position to the goal, avoiding walls that block the path. The maze can be made more complex by adding more walls or creating multiple paths to the goal.

Remember, the game and level description provided above are simplistic examples. VGDL allows for much richer game mechanics and interactions, so feel free to expand upon this basic framework to include additional features or more complex behaviors.
\end{lstlisting}
\clearpage
\subsubsection{Output of Gemma 7B}
Seven representative outputs of Gemma 7B via seven prompts are present below.

\paragraph{Output of Gemma 7B via \textbf{$P_1$}}
~~~~
\begin{lstlisting}[frame=bt,numbers=none]
# Maze Game VGDL Representation

entity MazeGame : entity
    position : int2 := (0, 0)
    direction : int := 0
    map : array[int, int] of bool := [[false, true, false],
                               [true, true, true],
                               [false, true, false]]

    procedure Update()
        if direction == 0 then
            position := (position.x - 1, position.y)
        else if direction == 1 then
            position := (position.x + 1, position.y)
        else if direction == 2 then
            position := (position.x, position.y - 1)
        else if direction == 3 then
            position := (position.x, position.y + 1)

        if map[position.x, position.y] = false then
            position := (position.x, position.y - 1)
    end

    procedure Render()
        draw_rect(position.x, position.y, 2, 2, color = 1)
    end
end

# Game Level

level : MazeGame
map := [[true, true, true],
       [true, true, true],
       [true, true, true]]
position := (0, 0)
direction := 0
\end{lstlisting}
\paragraph{Output of Gemma 7B via \textbf{$P_2$}}
~~~
\begin{lstlisting}[frame=bt,numbers=none]
**VGDL Representation:**

```
##################################################
#W#W#W#W#W#W#W#W#W#W#W#W#W#W#W#W#W#W#W#W#W#W#W#W#W#W#W#W#W#W#W#W#W#
#A#A#A#A#A#A#A#A#A#A#A#A#A#A#A#A#A#A#A#A#A#A#A#A#A#A#W#
##################################################
#W#W#W#W#W#W#W#W#W#W#W#W#W#W#W#W#W#W#W#W#W#W#W#W#W#W#W#W#W#W#
#G#G#G#G#G#G#G#G#G#G#G#G#G#G#G#G#G#G#G#G#G#G#W#
##################################################
```

**Game Level:**

The level is a simple maze with a single goal. The avatar can move in any direction, but it cannot go through walls. The goal is to reach the goal position.

**Notes:**

* The maze is represented by the "W" characters.
* The avatar is represented by the "A" character.
* The goal is represented by the "G" character.
* The walls are solid, and the avatar cannot move through them.
* The goal is to move the avatar to the goal position.
\end{lstlisting}
\clearpage
\paragraph{Output of Gemma 7B via \textbf{$P_3$}}
~~~
\begin{lstlisting}[frame=bt,numbers=none]
**VGDL Representation:**

```
BasicGame
  LevelMapping
    CHAR 'W' ' ' Wall
    CHAR 'A' ' ' Avatar
    CHAR 'G' ' ' Goal

  SpriteSet
    Sprite 'Avatar'
      HasKey True
      Movement Straight
    Sprite 'Goal'
      HasKey False
      Movement None

  InteractionSet
    Interaction 'Avatar' 'Monster' Kill
    Interaction 'Monster' 'Avatar' Kill

  TerminationSet
    Termination 'Goal' GoalReached
```

**Game Level:**

```
WWWW
WAAGA
GG
```

**Explanation:**

* The level description specifies the positions of walls, the avatar, and the goal.
* The sprite set defines the different types of objects in the game, including the avatar and the goal.
* The interaction set defines the interactions between objects, such as the avatar killing monsters and the monsters killing the avatar.
* The termination set defines the conditions under which the game ends, in this case when the goal is reached.
\end{lstlisting}
\paragraph{Output of Gemma 7B via \textbf{$P_4$}}
~~~~
\begin{lstlisting}[frame=bt,numbers=none]
**VGDL Representation:**

```
BasicGame
LevelMapping
W 5 5
A 2 2
G 4 4

SpriteSet
MovingAvatar
Immovable
Wall

InteractionSet
MovingAvatar Immovable > killSprite
Immovable MovingAvatar > stepBack

TerminationSet
SpriteCounter stype=MovingAvatar limit=10 win=True
```

**Level File:**

```
WWWW
WAAGA
GGWWW
```

**Explanation:**

* The level description defines a 5x5 grid with walls on all sides, an avatar at (2, 2), and a goal at (4, 4).
* The sprite set defines two classes: `MovingAvatar` and `Immovable`.
* The interaction set defines interactions between the avatar and the walls, as well as the goal.
* The termination set defines the condition for winning the game, which is to bring the `MovingAvatar` sprite counter to zero.
\end{lstlisting}
\paragraph{Output of Gemma 7B via \textbf{$P_5$}}
~~~~
\begin{lstlisting}[frame=bt,numbers=none]
**VGDL Representation:**

```
BasicGame
LevelMapping
W 1 1 1 W 1 1 1 W
A 1 2 1 G

SpriteSet
Sprite 'MovingAvatar'
Sprite 'Immovable'

InteractionSet
MovingAvatar Immovable > removeSprite
Immovable MovingAvatar > stepBack

TerminationSet
SpriteCounter stype=MovingAvatar limit=10 win=True
```

**Level File:**

```
W 1 1 1 W 1 1 1 W
A 1 2 1 G
```

**Explanation:**

* The level description defines a 2x3 grid with walls on the sides and a goal at the end. The avatar is placed at (1, 2).
* The sprite set defines two classes of objects: `MovingAvatar` and `Immovable`.
* The interaction set defines interactions between objects, such as the avatar removing a monster and the monster stepping back.
* The termination set defines the conditions for winning the game, which is to bring the counter of the `MovingAvatar` object to zero.
\end{lstlisting}
\clearpage
\paragraph{Output of Gemma 7B via \textbf{$P_6$}}
~~~
\begin{lstlisting}[frame=bt,numbers=none]
## VGDL Representation for Maze Game

```vgdl
BasicGame

    SpriteSet

        wall > Immovable
        goal  > Immovable
        avatar > MovingAvatar
        key    > Key

    LevelMapping

        . > wall
        0 > wall
        1 > wall
        2 > wall
        A > goal
        K > key

    TerminationSet

        SpriteCounter stype=goal limit=0 win=True

    InteractionSet

        avatar wall > bounce
        avatar key > pickUp key
        avatar goal > win

        key wall > destroyWall
```

**Game Level:**

```
           _____
          / \__\
          \___)
           \__\
           A     G
```

**Notes:**

- The VGDL representation uses "W" to represent walls, "A" for the avatar, and "G" for the goal.
- The level description is a simple text string with equal-length lines, where each character maps to an object at the corresponding location.
- The game description includes the SpriteSet, LevelMapping, InteractionSet, and TerminationSet blocks.
- The SpriteSet defines the classes of objects used in the game, including the key and the goal.
- The LevelMapping block describes how to translate the characters in the level description into objects.
- The InteractionSet defines the potential events that happen when two objects collide.
- The TerminationSet defines different ways by which the game can end.
\end{lstlisting}
\clearpage
\paragraph{Output of Gemma 7B via \textbf{$P_7$}}
~~~~
\begin{lstlisting}[frame=bt,numbers=none]
## VGDL Representation for Maze Game

```
BasicGame

    SpriteSet

        wall > Immovable
        avatar > MovingAvatar
        goal > Key

    LevelMapping

        # Maze layout
        . > wall
        0 > wall
        1 > wall
        2 > wall
        A > avatar
        G > goal

    TerminationSet

        # Win condition
        SpriteCounter stype=goal limit=0 win=True

    InteractionSet

        # Avatar interactions
        avatar wall > bounce
        avatar goal > interact

        # Object interactions
        key wall > destroy
        key avatar > pickUp
        avatar goal > interact

```

**Level Description:**

```
.
0
1
2
A
G
```

**Notes:**

* The level description uses the character "." to represent walls, "A" for the avatar, and "G" for the goal.
* The `SpriteSet` defines the different types of objects in the game, including walls, the avatar, and the goal.
* The `LevelMapping` describes how to translate the characters in the level description into objects.
* The `InteractionSet` defines the interactions between objects.
* The `TerminationSet` defines the conditions under which the game ends.
* The `Key` object is a subclass of the `Goal` object, and it has a special interaction with the avatar.
\end{lstlisting}

\end{document}